\documentclass[conference]{IEEEtran}
\IEEEoverridecommandlockouts
\usepackage{cite}
\usepackage{amsmath,amssymb,amsfonts}
\usepackage{algorithmic}
\usepackage{graphicx}
\usepackage{textcomp}
\usepackage{xcolor}
\usepackage{booktabs}
\usepackage{multirow, makecell}
\def\BibTeX{{\rm B\kern-.05em{\sc i\kern-.025em b}\kern-.08em
    T\kern-.1667em\lower.7ex\hbox{E}\kern-.125emX}}
\begin{document}

\title{MTLSI-Net: A Linear Semantic Interaction Network for Parameter-Efficient Multi-Task Dense Prediction}

\author{
\IEEEauthorblockN{Chen Liu$^{1}$, Hengyu Man$^{1 \star}$, Xiaopeng Fan$^{1,2,3}$, Debin Zhao$^{1}$}
\IEEEauthorblockA{$^{1}$Harbin Institute of Technology, $^{2}$Peng Cheng Laboratory, $^{3}$Harbin Institute of Technology SuZhou Research Institute}
\thanks{© 2026 IEEE. Personal use of this material is permitted. Permission from IEEE must be obtained for all other uses, in any current or future media, including reprinting/republishing this material for advertising or promotional purposes, creating new collective works, for resale or redistribution to servers or lists, or reuse of any copyrighted component of this work in other works.}
}

\maketitle

\begin{abstract}

Multi-task dense prediction aims to perform multiple pixel-level tasks simultaneously. However, capturing global cross-task interactions remains non-trivial due to the quadratic complexity of standard self-attention on high-resolution features. To address this limitation, we propose a Multi-Task Linear Semantic Interaction Network (MTLSI-Net), which facilitates cross-task interaction through linear attention. Specifically, MTLSI-Net incorporates three key components: a Multi-Task Multi-scale Query Linear Fusion Block, which captures cross-task dependencies across multiple scales with linear complexity using a shared global context matrix; a Semantic Token Distiller that compresses redundant features into compact semantic tokens, distilling essential cross-task knowledge; and a Cross-Window Integrated attention Block that injects global semantics into local features via a dual-branch architecture, preserving both global consistency and spatial precision. These components collectively enable the network to capture comprehensive cross-task interactions at linear complexity with reduced parameters. Extensive experiments on NYUDv2 and PASCAL-Context demonstrate that MTLSI-Net achieves state-of-the-art performance, validating its effectiveness and efficiency in multi-task learning.
\end{abstract}


\section{Introduction}

In computer vision, multi-task dense prediction \cite{survey_simon} has emerged as a paradigm for training a unified model that simultaneously tackles multiple pixel-level tasks, such as semantic segmentation, depth estimation, and surface normal estimation. Compared with separate single-task models, multi-task learning (MTL) offers significant advantages in both efficiency and performance. The unified model substantially reduces parameter redundancy and computational overhead, while enabling different tasks to benefit from shared representations and cross-task knowledge transfer. While techniques such as meta learning mechanisms \cite{man2024metaip}, data-driven structured modeling \cite{10155660}, and graph-based relational reasoning \cite{li2025langgraph, li2025svgri} have advanced neural network design for individual visual tasks, the key challenge in multi-task learning lies in how to effectively model cross-task interactions.

Cross-task interaction \cite{Cross-Stitch, AutoMTL, MTL-NAS, taskprompter,PAP-Net, MTI-Net, PSD, InvPT++, TIF} plays a crucial role in multi-task learning for capturing global dependencies and relationships across tasks. Early approaches model cross-task relationships with Convolutional Neural Networks (CNNs) \cite{Cross-Stitch, PAD-Net, PAP-Net} or Multilayer Perceptrons (MLPs) \cite{MTFormer} for each task. However, their performance is inherently constrained by the local receptive context, limiting the capture of long-range cross-task dependencies. The advent of Transformers introduces Multi-Head Self-Attention (MHSA) \cite{Transformer}, which offers a powerful mechanism for modeling global relationships. DeMT \cite{DeMT} directly applies MHSA to capture interactions across $T$ task features. However, dense prediction tasks typically operate on high-resolution feature maps, containing tens of thousands of tokens, which makes the $\mathcal{O}(N^2)$ complexity of standard attention computationally prohibitive. To address this challenge, existing methods adopt different strategies. Some prior works \cite{InvPT++, InvPT} first decrease feature resolutions to establish spatial and cross-task interaction via attention computation, then progressively upsample the multi-task features at different scales to save complexity. While this makes global attention feasible, performing interaction on low-resolution features inevitably leads to the loss of spatial details. Other approaches utilize a limited number of learnable parameters to aggregate cross-task context. For instance, TaskPrompter \cite{taskprompter} employs prompts to learn task-specific representations, and models the cross-task relationships through the interactions between task prompts. Similarly, the cross-task query module in \cite{MQTransformer} focus on query-level interactions, thereby reducing complexity. While these methods are efficient, the sparse information flow hinders the modeling of dense pixel-to-pixel correlations. Another line of work \cite{TIF} explores the features at lower scales with MHSA, and propagates global relationships from low-level to high-level scales. Despite these efforts, existing methods still rely on quadratic attention mechanisms and ultimately sacrifice either spatial resolution or the density of cross-task interaction. This underscores the need for a lightweight yet effective approach to cross-task interaction.

Linear attention \cite{linear,EfficientViT,ahn2024linear} offers a theoretically attractive alternative by replacing softmax normalization with kernel-based approximations, achieving linear complexity with respect to sequence length in segmentation \cite{EfficientSAM} and compression \cite{DCA} tasks. 
In a similar spirit, other sub-quadratic sequence models such as RWKV \cite{liu2026t} have demonstrated strong capability in learning compact visual representations, while bio-inspired spiking architectures \cite{li2025stft} and efficient generative frameworks \cite{wang2026t} further validate the broad potential of lightweight neural designs.
However, existing linear attention methods are primarily designed for single-task scenarios, limiting their ability to leverage cross-task semantic dependencies on multi-task learning.

To address the aforementioned challenges, we propose a novel framework named \textbf{M}ulti-\textbf{T}ask \textbf{L}inear \textbf{S}emantic \textbf{I}nteraction \textbf{Net}work (MTLSI-Net), which preserves dense cross-task semantic interaction while enriching task-specific features with cross-task semantic information. To capture comprehensive cross-task semantic dependencies without sacrificing spatial resolution or interaction density, we introduce the Multi-Task Multi-scale Query Linear Fusion Block (MT-MQLFB), which employs an efficient linear attention mechanism architecture. Specifically, MT-MQLFB uses convolutional kernels of different sizes to generate multi-scale queries across tasks and constructs a shared global context matrix. 
% This allows it to capture cross-task semantic dependencies with different receptive fields at linear complexity.
Although MT-MQLFB leverages linear attention mechanism to model cross-task dependencies, high-resolution multi-task features inevitably contain spatial redundancy. Therefore, we introduce a Semantic Token Distiller module to extract the effective cross-task context. By reducing sequence length, the Distiller adaptively aggregates the fused sequence into compact semantic tokens, distilling the most relevant cross-task semantic dependencies. To ensure these high-level semantics effectively guide the dense prediction tasks, we propose the \textbf{C}ross-\textbf{W}indow \textbf{I}nteraction \textbf{B}lock (CWIB) that bridges the gap between global semantics and local details via a dual-branch architecture. Specifically, CWIB explicitly injects the distilled semantic tokens into local features through cross-attention for global guidance, while maintaining local spatial coherence via window-based self-attention. Extensive experiments on NYUDv2 \cite{NYUD} and PASCAL-Context \cite{pascal} demonstrate that our method achieves state-of-the-art performance while maintaining superior parameter efficiency.

\section{Method}
\begin{figure*}[t]
    \centering
    \includegraphics[width=0.9\textwidth]{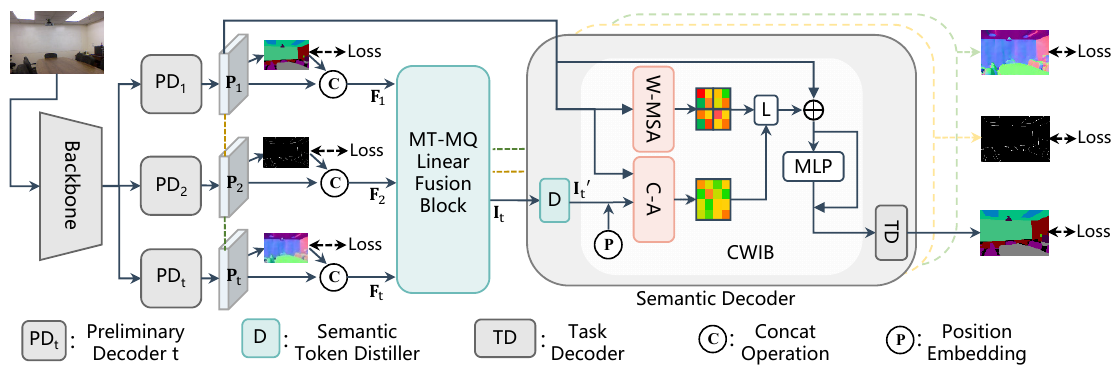}
    \caption{Overview of the proposed coarse-to-fine multi-task learning framework. Given an input image, a shared backbone first extracts multi-scale features. Subsequently, preliminary decoders generate initial task-specific representations and coarse predictions. Next, to enable cross-task interaction, the MT-MQLFB processes the concatenated features and predictions. Finally, semantic decoders, incorporating the Semantic Token Distiller and Cross-Window Integrated Attention Block (CWIB), refine these features to produce fine-grained predictions for all tasks.}
    \label{fig:framework}
\end{figure*}

\subsection{Framework Overview}

As illustrated in Fig. \ref{fig:framework}, the proposed MTLSI-Net follows a coarse-to-fine paradigm \cite{PAD-Net, InvPT}, and primarily consists of three main components: a shared backbone with preliminary decoders, a Multi-task Multi-scale Query Linear Fusion Block (MT-MQLFB), and semantic decoders. Initially, the shared backbone extracts multi-scale features for all tasks. Subsequently, preliminary decoders generate initial task-specific representations $(\mathbf{P}_1,...,\mathbf{P}_t)$ from these features and produce corresponding coarse predictions. However, these coarse predictions often suffer from suboptimal performance due to the lack of effective cross-task interactions. To mitigate this limitation, the MT-MQLFB is designed to capture cross-task semantic dependencies across different tasks. Specifically, for each task, the input feature $\mathbf{F}_i$ is obtained by concatenating  $\mathbf{P}_i$ with its corresponding coarse prediction, and is then projected to a unified size $(H, W, d)$. Finally, the semantic decoders, which consist of the Semantic Token Distiller and Cross-Window Integrated attention Block (CWIB), enhance the initial task-specific features with the cross-task semantics captured by MT-MQLFB, producing pixel-level predictions for all tasks. Detailed descriptions of these modules are provided in the following sections.

\begin{figure}[t]
    \centering
    \includegraphics[width=1\columnwidth]{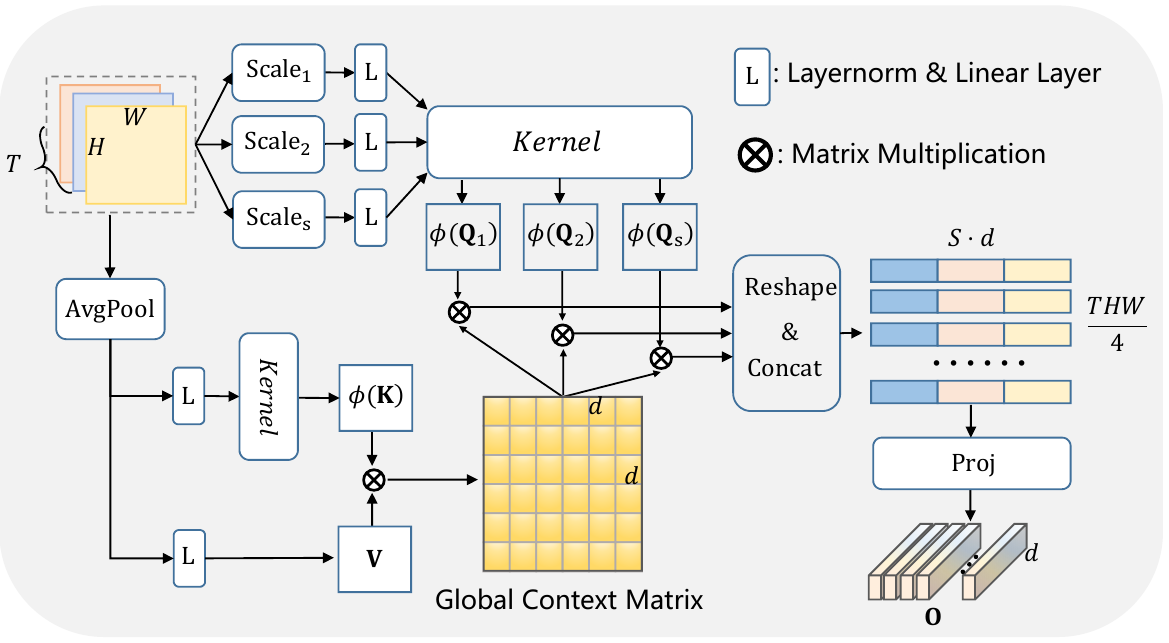}
    \caption{Architecture of the Multi-task Multi-scale Query Linear Fusion Block.}
    \label{fig:lfa}
\end{figure}

\subsection{Multi-task Multi-scale Query Linear Fusion Block}
\label{sec:2b}

Existing multi-task methods typically incorporate specialized self-attention modules to model diverse cross-task dependencies. However, the quadratic complexity incurs prohibitive computational overhead when modeling cross-task interactions on high-resolution feature maps. To overcome this limitation, we propose the Multi-Task Multi-scale Query Linear Fusion Block (MT-MQLFB). MT-MQLFB leverages the $\mathcal{O}(N)$ complexity of the linear attention mechanism to efficiently fuse task-specific features across multiple scales, while simultaneously capturing rich cross-task semantic interaction information.

In contrast to standard self-attention, linear attention \cite{linear} substitutes the softmax normalization with a kernel-based approximation and alters the matrix multiplication order, thereby achieving linear complexity with respect to sequence length:

\begin{equation}
\mathbf{O}=\frac{\phi(\mathbf{Q})\left(\phi(\mathbf{K})^{\top}\mathbf{V}\right)}{\phi(\mathbf{Q})\sum_j\phi(\mathbf{K})_j^{\top}},
\label{eq:1}
\end{equation}
where $\phi(\cdot)$ denotes a kernel function that ensures non-negative outputs. However, directly applying (\ref{eq:1}) to multi-task learning is suboptimal, as it cannot model the complex cross-task interactions with task-specific receptive field requirements. In multi-task dense prediction, different tasks inherently prefer different receptive fields \cite{MTI-Net, hrnet}. For instance, edge detection emphasizes fine-grained local patterns, while semantic segmentation relies on broader regional context. Consequently, we extend the kernelized approximation $\phi(\mathbf{Q})\phi(\mathbf{K})^{\top}$ to multi-task scenarios with multi-scale queries.

As depicted in Fig. \ref{fig:lfa}, to capture cross-task semantic dependencies at different granularities, we generate multi-scale queries for each task using convolutional kernels of varying sizes in $\mathrm{Scale}_s$. Specifically, for scale $s$ we construct features $\mathbf{G}_s$ :
\begin{equation}
    \mathbf{G}_s=\mathrm{Concat}(\{\mathrm{Re}(\mathrm{Conv}^{s\times s}(\mathbf{F}_{i}))\}_{i=1}^{T}), s\in \{1,3,5\},
    \label{eq:2}
\end{equation}
where $\mathbf{F}_i \in \mathbb{R}^{H \times W \times d}$ denotes the input features for the $i$-th task. $\mathrm{Conv}^{s\times s}$ represents a composite block consisting of a depth-wise convolution with kernel size $s$, followed by Batch Normalization and ReLU activation. $\mathrm{Re}(\cdot)$ denotes the reshaping of spatial dimensions from $(d, H/2, W/2)$ into a sequence of tokens of size $(HW/4, d)$. The processed features across all tasks at specific scale $s$ are then concatenated along the token dimension. The query $\mathbf{Q}_s$ is projected from $\mathbf{G}_s$ through a linear layer, utilizing $\text{ELU}(\cdot) + 1$ as the kernel function. Simultaneously, to construct a shared global context, the keys $\mathbf{K}$ and values $\mathbf{V}$ are derived from the feature $\mathbf{M}$:
\begin{equation}
\begin{gathered}
    \mathbf{M} = \mathrm{Concat}(\{\mathrm{Re}(\mathrm{Pool}(\mathbf{F}_i))\}_{i=1}^T), \\
    \mathbf{K} = \mathrm{LN}(\mathbf{M})\mathbf{W}^{K}, 
    \mathbf{V} = \mathrm{LN}(\mathbf{M})\mathbf{W}^{V}, 
    \mathbf{Q_s} = \mathrm{LN}(\mathbf{G_s})\mathbf{W}^{Q}_{s},
\end{gathered}
    \label{eq:3}
\end{equation}
where $\mathrm{Pool}$ denotes 2D pooling and $\mathrm{LN}$ represents layer normalization. Following the linear attention mechanism, the output $\mathbf{O}_s$ is computed as (\ref{eq:4}):
\begin{equation}
    \begin{aligned}\mathbf{O}_{s}&=\frac{\phi\left(\mathbf{Q}_{s}\right)\left(\phi(\mathbf{K})^{\top}\mathbf{V}\right)}{\phi\left(\mathbf{Q}_{s}\right)\sum_{j}\phi(\mathbf{K})_{j}^{\top}}\\&=\frac{\phi\left(\mathbf{W}_{s}^{Q}\mathbf{G}_{s}\right)\left(\phi(\mathbf{W}^{K}\mathbf{M})^{\top}\mathbf{W}^{V}\mathbf{M}\right)}{\phi\left(\mathbf{W}_{s}^{Q}\mathbf{G}_{s}\right)\sum_{j}\phi(\mathbf{W}^{K}\mathbf{M})_{j}^{\top}}.\end{aligned}
    \label{eq:4}
\end{equation}
Note that the term $\frac{\phi(\mathbf{W}^{K}\mathbf{M})^{\top}\mathbf{W}^{V}\mathbf{M}}{\sum_{j}\phi(\mathbf{W}^{K}\mathbf{M})_{j}^{\top}}$ in (\ref{eq:4}), which represents global context matrix, is independent of the queries $\mathbf{Q}_s$. Consequently, it needs to be computed only once and is shared across different scales. Finally, the outputs $\mathbf{O_s}$ from different scales are projected to the same dimensional space and aggregated via MLP to generate final fused features $\mathbf{I}$:
\begin{equation}
\begin{aligned}
    \mathbf{O}_{agg} &= \mathrm{Concat}(\mathbf{O}_1, ..., \mathbf{O}_s)\mathbf{W}^O, \\
    \mathbf{I} &= \mathbf{O}_{agg} + \mathrm{MLP}(\mathbf{O}_{agg}).
\end{aligned}
    \label{eq:5}
\end{equation}

In the $T$ dense prediction tasks with $S$ scales setting, the computational complexity of MT-MQLFB is linear with respect to the feature resolution $(H, W)$, specifically $\mathcal{O}(S \cdot \frac{THW}{4} \cdot d^2)$. Given that $ST \ll HW$, the complexity is substantially lower than that of the standard multi-head attention, which incurs a quadratic complexity of $\mathcal{O}((THW)^2 \cdot d)$ \cite{DeMT, InvPT, InvPT++, TaskExpert}. The linear efficiency allows MT-MQLFB to process more tokens across tasks and scales, thereby facilitating more comprehensive cross-task interactions for dense prediction.

\subsection{Semantic Token Distiller}

% To achieve fine-grained predictions, we employ task-specific semantic decoders to improve the preliminary representations $\mathbf{P_t}$ for task $t$. 
Although fused feature $\mathbf{I}$ encapsulates the required cross-task interactions, the feature aggregation inevitably contains task-irrelevant components from other tasks that interferes with task-specific predictions. Therefore, we introduce a Semantic Token Distiller to eliminate the redundancy while preserving task-relevant cross-task dependencies. Additionally, the distillation compresses the sequence length, thereby alleviating computation and memory overhead in the subsequent decoding process. 
Distinct from knowledge distillation involving teacher-student models, our module distills the token sequence by condensing $\mathbf{I}$ into a compact set of representative semantic tokens via learned soft assignment.

Specifically, the task-specific Semantic Token Distiller learns a compact representation for each task $t$, mapping the fused feature $\mathbf{I}$ into $K$ representative tokens with $K \ll N$. We first employ a lightweight MLP, composed of batch normalization and 1D convolutions, to project $\mathbf{I}$ into a semantic assignment matrix $\mathbf{A_t} \in \mathbb{R}^{K \times N}$. A Softmax function is then applied along the token dimension of $\mathbf{A_t}$ to normalize the weights, characterizing the contribution of all original tokens to the $K$ semantic tokens. Finally, the distilled features $\mathbf{I}_t' \in \mathbb{R}^{K \times d}$ are generated by computing the weighted sum of the linearly projected $\mathbf{I}$ according to $\mathbf{A_t}$:
\begin{equation}
\begin{aligned}
    \mathbf{A_t} = \text{Softmax}(\mathrm{MLP_t}(\mathbf{I}^{\top})) \in \mathbb{R}^{K \times N}, \\
    \mathbf{I}_t' = \mathbf{A_t} \cdot \mathrm{Proj_t}(\mathbf{I}) \in \mathbb{R}^{K \times d},
\end{aligned}
\end{equation}
where $\mathrm{Proj_t}$ denotes a linear layer and $d$ represents the feature dimension.

\subsection{Cross-Window Integrated attention Block}

Preliminary features preserve local structural patterns, while the absence of cross-task context in MTL constrains their representational capability. To address this limitation, we propose the Cross-Window Integrated Attention Block (CWIB), which explicitly incorporates cross-task semantic guidance $\mathbf{I}_t'$, derived from the Semantic Token Distiller, into the preliminary features $\mathbf{P}_t$. 
% The CWIB employs a dual-branch attention structure that integrates cross-task interactions with local spatial dependencies. 
For brevity, we omit the task subscript $t$ in the following formulations, using $\mathbf{I}'$ and $\mathbf{P}$ to denote $\mathbf{I}_t'$ and $\mathbf{P}_t$, respectively.

The CWIB consists of two parallel branches: a Window-based Self-Attention (W-MSA) branch to preserve spatial coherence with low computational complexity, and a Cross-Attention (CA) branch to inject semantic information. For the W-MSA, the feature $\mathbf{P}$ is projected into query $\mathbf{Q}'$, key $\mathbf{K}_w$ and value $\mathbf{V}_w$ as:
\begin{equation}
    \mathbf{Q}'=\mathrm{LN}(\mathbf{P})\mathbf{W}',
    \mathbf{K}_w=\mathrm{LN}(\mathbf{P})\mathbf{W}_w^K,
    \mathbf{V}_w=\mathrm{LN}(\mathbf{P})\mathbf{W}_w^V,
    \label{eq:7}
\end{equation}
where $\mathrm{LN}$ denotes layer normalization. The window attention $\mathbf{O}_w$ is then computed as:
\begin{equation}
    \mathbf{O}_w = \mathrm{W}\text{-}\mathrm{MSA}(\mathbf{Q}', \mathbf{K}_w, \mathbf{V}_w),
    \label{eq:8}
\end{equation}
in which standard non-overlapping windows of size $(H_w, W_w)$ are employed. For cross-attention, the key $\mathbf{K_c}$ and the value $\mathbf{V_c}$ are projected from the cross-task semantic $\mathbf{I'}$ with task-specific learnable positional embedding $\mathbf{E}$:
\begin{equation}
    \mathbf{K}_c = \mathrm{LN}(\mathbf{I}'+\mathbf{E})\mathbf{W}_c^K, \mathbf{V}_c=\mathrm{LN}(\mathbf{I}'+\mathbf{E})\mathbf{W}_c^V.
\end{equation}
The cross attention $\mathbf{O}_c$ is derived via Multi-Head Attention:
\begin{equation}
    \mathbf{O}_c = \mathrm{CA}(\mathbf{Q}', \mathbf{K}_c, \mathbf{V}_c).
\end{equation}
Subsequently, the outputs from both branches are concatenated and linearly projected, followed by a residual connection. This integration complements the local spatial coherence preserved by $\mathrm{W}\text{-}\mathrm{MSA}$ with the cross-task semantic context captured by $\mathrm{CA}$, yielding representations that incorporate both fine-grained spatial structures and beneficial cross-task dependencies. Following standard Transformer design, a Feed-Forward Network (FFN) is employed to produce the integrated features $\mathbf{X}$:
\begin{equation}
\begin{aligned}
    \mathbf{Z} &= \mathrm{Linear}(\mathrm{Concat}(\mathbf{O}_w, \mathbf{O}_c)) + \mathbf{P}, \\
    \mathbf{X} &= \mathbf{Z} + \mathrm{FFN}(\mathrm{LN}(\mathbf{Z})).
\end{aligned}
\end{equation}

The complexity of the proposed attention mechanism is $\mathcal{O}(HW(H_wW_w+K)\cdot d)$, linear with respect to the task-specific feature map size. The integrated features $\mathbf{X}$ are then passed to a task-specific output decoder, structurally similar to the preliminary decoder, to generate the final predictions.

\section{Experiments}

\begin{table*}[ht]
\centering
\caption{Performance comparison with SOTA methods using different backbones on NYUDv2.}
\label{table:1}
\begin{tabular}{c|c|ccccc}
\toprule
Method        & Backbone & Seg.(mIoU$\uparrow$) & Depth(RMSE$\downarrow$) & Normal(mErr$\downarrow$) & Boundary(odsF$\uparrow$) & Params(M) \\ \midrule
PAD-Net \cite{PAD-Net}       & HRNet-18  & 36.70      & 0.6264      & 20.85        & \textbf{76.50 }     &5.02        \\
PSD \cite{PSD}           & HRNet-18  & 36.69      & 0.6246      & 20.87        & \underline{76.42}      &4.71        \\
ATRC \cite{ATRC}          & HRNet-18  & \underline{38.90}      & \underline{0.6010}      & \underline{20.48}        & 76.34      &5.06        \\ 
Ours          & HRNet-18  & \textbf{42.89}      & \textbf{0.5892}      & \textbf{20.12}       & 76.40     &13.94            \\\midrule
MQTransformer \cite{MQTransformer} & Swin-T   & 43.61      & 0.5979      & \underline{20.05}        & 76.20      &32.91       \\
DeMT \cite{DeMT}          & Swin-T   & 46.36      & 0.5871      & 20.65        & \textbf{76.90}      &32.07       \\
InvPT++ \cite{InvPT++}       & Swin-T   & 44.94      & \textbf{0.5554}      & 20.43        & 76.20      &-   \\
Ours         & Swin-T   & \textbf{46.79}      & \underline{0.5766}      & \textbf{19.82}        & \underline{76.60}      &17.45    \\ \midrule
InvPT \cite{InvPT}         & ViT-L    & 53.56      & 0.5183      & 19.04        & 78.10      & 97.96    \\
TaskPrompter \cite{taskprompter}  & ViT-L    & 55.30      & 0.5152      & \underline{18.47}        & 78.20      & 88.00    \\
TaskExpert \cite{TaskExpert}    & ViT-L    & 55.35      & 0.5157      & 18.54        & 78.40      & -    \\
TIF \cite{TIF}           & Swin-L   & \underline{56.80}      & \underline{0.5023}      & 19.21        & \textbf{79.52}      & 32.91     \\
Ours          & Swin-L   & \textbf{57.22}      & \textbf{0.4904}      & \textbf{18.26}        & \underline{78.60}      & 38.27    \\ \bottomrule
\end{tabular}
\end{table*}

\subsection{Experimental Setup}
\textbf{Datasets and Metrics}. We evaluate our method on two multi-task dense prediction benchmarks: NYUDv2 \cite{NYUD} and PASCAL-Context \cite{pascal}. NYUDv2 comprises four tasks: semantic segmentation (Seg.), monocular depth estimation (Depth), surface normal estimation (Normal), and object boundary detection (Boundary). PASCAL-Context includes Seg., Boundary, Normal, human parsing (Parsing), and saliency detection (Sal.). We adopt mean Intersection over Union (mIoU) for Seg. and Parsing, Root Mean Square Error (RMSE) for Depth, mean Error (mErr) for Normal, maximal F-measure (maxF) for Sal. and optimal-dataset-scale F-measure (odsF) for Boundary.

\textbf{Implementation Details}. Following \cite{ATRC, DeMT}, input images are resized to $448\times576$ for NYUDv2 and $512\times512$ for PASCAL-Context. We employ the $\mathrm{\ell_1}$ loss for both Depth and Normal, and apply cross-entropy loss for the remaining tasks. The loss weights for each task are consistent with \cite{ATRC}. We utilize the Swin Transformers \cite{swin}, pre-trained on ImageNet-22K \cite{imagenet}, as the backbones. The multi-scale queries use kernel size $\{1,3,5\}$. The window size is set to $14 \times 16$ in CWIB on NYUDv2 and $16 \times 16$ on PASCAL-Context. The model is optimized using the AdamW optimizer with a polynomial learning rate scheduler, starting with an initial learning rate of $2\times10^{-5}$. Training is conducted for 40,000 iterations, with batch sizes of 4 and 8 for PASCAL-Context and NYUDv2, respectively.

\subsection{Comparison with State-of-the-art Methods}

In this section, we compare the performance of MTLSI-Net with recent state-of-the-art approaches on the NYUDv2 and PASCAL-Context benchmarks.

\textbf{Results on NYUDv2.} As shown in Table \ref{table:1}, ``$\uparrow$" denotes higher values are better, while ``$\downarrow$" indicates lower values are better. The reported parameter counts exclude the backbone network. We compare our method against both CNN-based and Transformer-based approaches across different backbone architectures. With the HRNet18 backbone, our method achieves substantial improvements over CNN-based methods, surpassing ATRC by 3.99 mIoU on Seg. and improving by 0.36 mErr on Normal. When using Swin-T as the backbone, our method outperforms InvPT++ by 1.85 mIoU on Seg. and 0.61 on Normal, while achieving the second-best performance on Depth and Boundary. Notably, compared to DeMT, which captures cross-task dependencies with standard self-attention, our method achieves superior performance on most tasks while using only approximately 54\% of the parameters (17.45M vs. 32.07M). With the Swin-L backbone, our method also achieves state-of-the-art performance with 38.27M parameters, demonstrating parameter efficiency.

\textbf{Results on PASCAL-Context.} Table \ref{table:2} reports the results obtained with large Transformer backbones across all methods. Our method achieves the best performance on Seg. (80.86 mIoU) and Parsing (69.90 mIoU), and the second-best on Normal (13.71 mErr), with a marginal gap of 0.15 from the best. Although slightly behind MQTransformer and InvPT++ on Boundary, our method maintains superior overall performance. While our FLOPs are higher due to operating on high-resolution intermediate features, MTLSI-Net uses 44\% fewer parameters (234M) than TaskExpert (420M) and remains more parameter-efficient than other MTL approaches while achieving similar or better accuracy. This demonstrates the parameter efficiency of our multi-task dense prediction framework.

\begin{table}[t]
\centering
\caption{Performance comparison with SOTA methods on PASCAL-Context.}
\label{table:2}
\setlength{\tabcolsep}{0.5pt}
\begin{tabular}{c|ccccc|cc}
\toprule
\multirow{2}{*}{Method} & Seg.  & Parsing & Sal.  & Normal & Boundary  & Params  & FLOPs\\
                          & mIoU$\uparrow$  & mIoU $\uparrow$   & maxF$\uparrow$  & mErr$\downarrow$  & odsF$\uparrow$   & M & G  \\ \midrule
PAD-Net \cite{PAD-Net}       & 78.01 & 67.12   & 79.21 & 14.37  & 72.60    & 330    & 773\\
MTI-Net \cite{MTI-Net}       & 78.31 & 67.40   & \underline{84.75} & 14.67  & 73.00    & 851    & 774\\
ATRC \cite{ATRC}          & 77.11 & 66.84   & 81.20 & 14.23  & 72.10    & 340    & 871\\
MQTransformer \cite{MQTransformer} & 78.93 & 67.41   & 83.58 & 14.21  & \textbf{73.90}  & - & - \\ 
TaskExpert \cite{TaskExpert}   & \underline{80.64}    & \underline{69.42}    & \textbf{84.87}   & \textbf{13.56}  & 73.30 & 420 & 622 \\
InvPT++ \cite{InvPT++}       & 80.22 & 69.12   & 84.74 & 13.73  & \underline{73.50}    & 421    & 667\\
Ours          & \textbf{80.86} & \textbf{69.90}   & 84.52 & \underline{13.71}  & 73.40    & 234    & 772 \\ \bottomrule
\end{tabular}
\end{table}

\subsection{Effect of token count $K$ in Semantic Token Distiller}

\begin{figure}[t]
\centering
    \includegraphics[width=0.9\columnwidth]{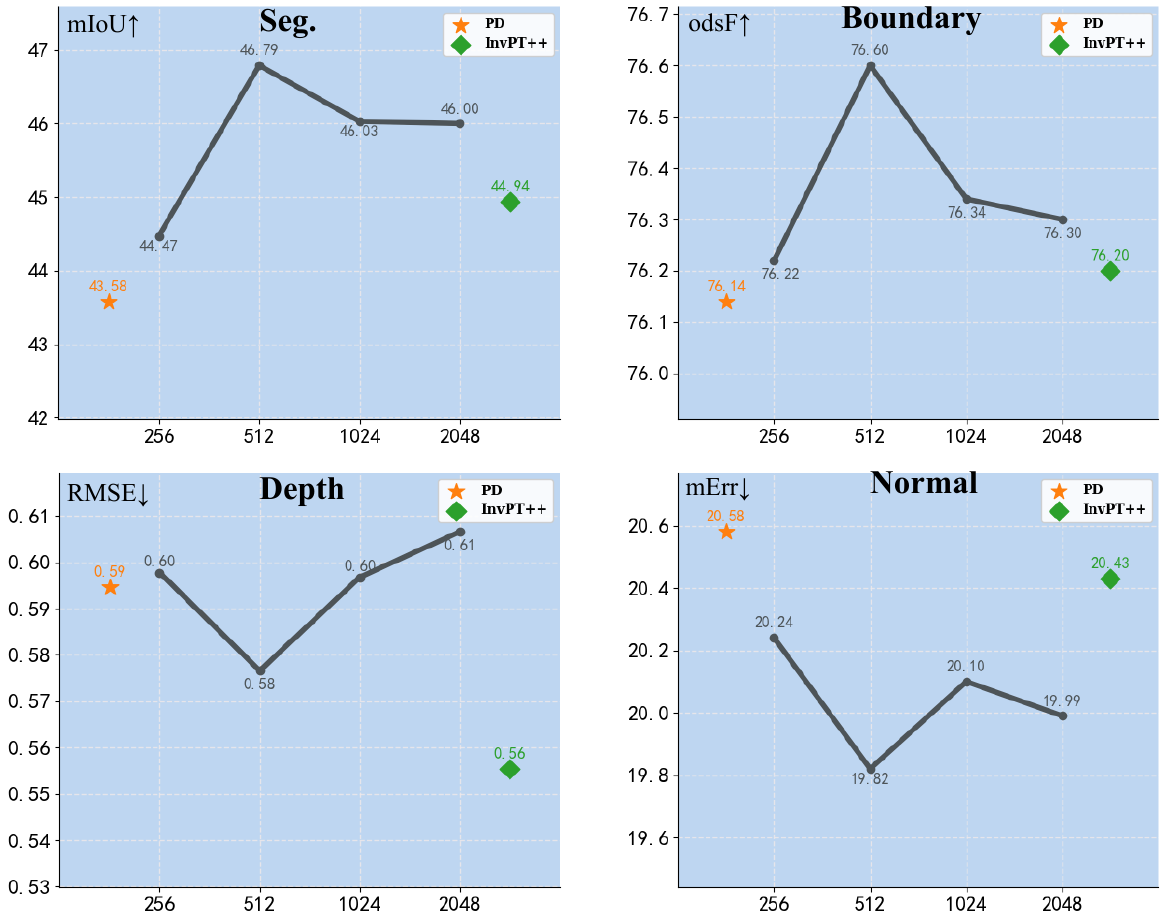}
\caption{Ablation study on the number of semantic tokens $K$ on NYUDv2.}
\label{fig:3}
\end{figure}

To obtain compact cross-task representations, the Semantic Token Distiller aggregates the integrated features $\mathbf{I}'$ with $K$ semantic tokens. We conduct ablation studies with the Swin-T backbone on the NYUDv2 dataset to investigate the impact of varying the token count $K$. Specifically, we evaluate $K \in \{256, 512, 1024, 2048\}$. For comparison, we establish baselines using only Preliminary Decoders (PD) with Swin-T and InvPT++.

As illustrated in Fig. \ref{fig:3}, our proposed method outperforms both PD and InvPT++ across most tasks for all values of $K$, with the sole exception of Depth, where the performance gap is marginal. This validates the effectiveness of the Semantic Token Distiller. Notably, a larger $K$ does not guarantee better performance. When $K$ increases from 512 to 2048, the Segmentation drops by 0.79 mIoU and Boundary decreases by 0.30 odsF. Among the four configurations, $K=512$ yields the best overall performance, achieving 46.79 mIoU on Seg., 19.82 mErr on Normal, and 76.60 odsF on Boundary. This observation suggests that while a sufficient number of tokens is necessary to capture semantic details, an excessive number may introduce redundancy into the semantic representation rather than contributing beneficial cross-task semantic dependencies.

\subsection{Ablation of multi-scale queries in MT-MQLFB}

\begin{table}[t]
\centering
\caption{Ablation study on the effect of multi-scale query configurations in MT-MQLFB.}
\label{table:3}
\begin{tabular}{c|c|cccc}
\toprule
\multirow{2}{*}{Dataset} & \multirow{2}{*}{scale} & Seg   & Depth  & Normal & Boundary \\
                         &                        & mIoU$\uparrow$    & RMSE$\downarrow$    &mErr$\downarrow$    &odsF$\uparrow$          \\ \midrule
\multirow{4}{*}{\makecell{NYUDv2\\($K=1024$)}}  & 1                      & 44.17 & 0.6254 & 20.51  & 76.00    \\
                         & 3                      & 44.10 & 0.5985 & 20.35  & 76.02    \\
                         & 5                      & 43.85 & 0.6014 & 20.34  & 75.96    \\
                         & (1,3,5)                & 46.03 & 0.5968 & 20.10  & 76.34    \\ \midrule
\end{tabular}
% \end{table}

% \begin{table}[]
\setlength{\tabcolsep}{3.5pt}
\centering
\begin{tabular}{c|c|ccccc}
\midrule
\multirow{2}{*}{Dataset}        & \multirow{2}{*}{scale} & Seg. & Parsing & Sal. & Normal & Boundary \\ 
                                &                        & mIoU$\uparrow$  & mIoU $\uparrow$   & maxF$\uparrow$  & mErr$\downarrow$  & odsF$\uparrow$         \\ \midrule
\multirow{4}{*}{\makecell{PASCAL-\\Context\\($K=512$)}} & 1  & 71.62  &  57.54   &  83.84  &  15.34  &   71.92       \\
                                & 3                      &  71.52     &  58.93   &  84.23  & 14.76       &  72.00        \\
                                & 5                      &  71.38    &  58.96   &  84.00     &  14.64      & 72.01         \\
                                & (1,3,5)                & 72.96  & 60.63  & 84.34  & 14.37  &  72.26       \\ \bottomrule
\end{tabular}
\end{table}

To investigate the impact of multi-scale queries within our proposed MT-MQLFB, we conduct an ablation study on different scale configurations using a Swin-T backbone.  As described in section \ref{sec:2b}, our multi-task multi-scale query linear attention maintains queries with distinct receptive fields. We compare individual single-scale configurations $s \in \{1, 3, 5\}$ against the proposed multi-scale query strategy $s = \{1, 3, 5\}$. Notably, when $s=1$, the module degenerates to standard linear attention.

Table \ref{table:3} presents the quantitative comparisons. The results demonstrate that the multi-scale configuration yields superior performance across all tasks compared to any single-scale counterpart. Single-scale settings struggle to balance performance across tasks. For example, Scale 1 performs adequately on Seg (44.17 mIoU). but poorly on Depth (0.6254 RMSE) in NYUDv2 dataset, whereas Scale 5 favors Normal but compromises Seg. accuracy in the PASCAL-Context dataset. We attribute the success of the aggregated strategy to our multi-query attention. By treating different scales as distinct queries, the model can simultaneously capture cross-task interactions at varying receptive fields. Consequently, the combined configuration $\{1, 3, 5\}$ achieves the overall best results on both datasets, validating that fusing multi-scale queries is essential for cross-task representation in linear attention mechanism.

\subsection{Qualitative Study}
\begin{figure}[t]
\centering
    \includegraphics[width=1\columnwidth]{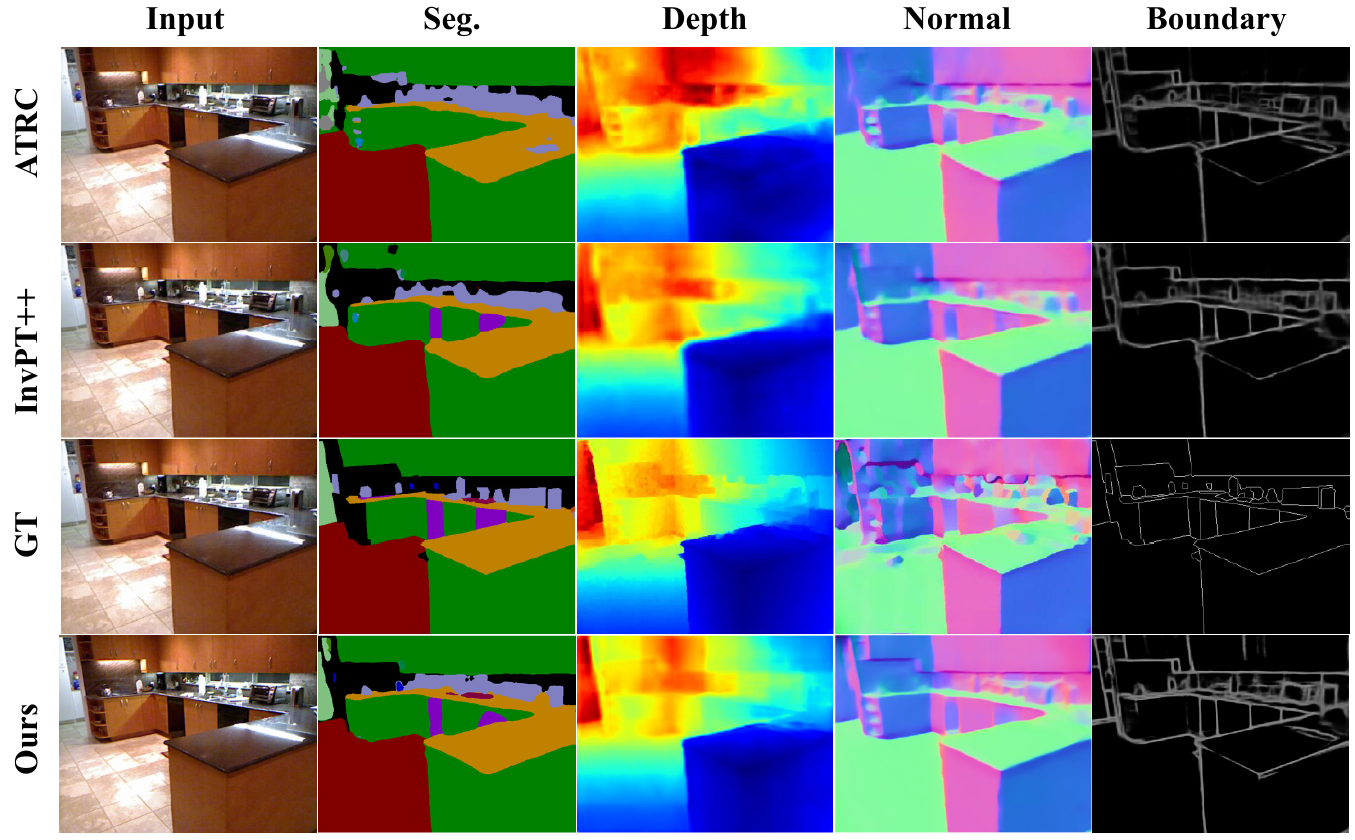}
\caption{Visual comparison on the NYUDv2 dataset.}
\label{fig:4}
\end{figure}

To better understand the performance of our MTLSI-Net, we provide qualitative visualizations of multi-task predictions in Fig. \ref{fig:4}. We compare our method, using Swin-L as the backbone, with ATRC and InvPT++ on the NYUDv2 dataset.

For Seg., our results exhibit sharper object boundaries and more precise object delineation. We attribute the improvement to the effective integration of the Boundary task, which provides explicit structural cues that refine segmentation borders. For Depth, MTLSI-Net accurately captures the details of nearby objects (typically shown in blue) while maintaining high consistency in the outlines of the distant field (shown in red). The Normal prediction shows smooth orientations within planar regions, and Boundary captures both salient and subtle edges. 

% Overall, the strong results demonstrate that our MTLSI-Net performs well in both cross-task information capture and interaction in multi-task dense predictions.

\section{Conclusion}

In this paper, we tackle the challenge of modeling global cross-task dependencies on high-resolution features for multi-task dense prediction. We present MTLSI-Net, a framework that shifts the paradigm from computationally expensive quadratic self-attention to a linear-complexity alternative without compromising representation capability. The effectiveness stems from three components: the MT-MQLFB captures multi-scale cross-task dependencies via linear attention; the Semantic Token Distiller eliminates spatial redundancy to form compact representations; and the CWIB seamlessly integrates these global semantics with local features. Comprehensive evaluations demonstrate that MTLSI-Net achieves state-of-the-art results with superior parameter efficiency. Future work will explore extending this linear interaction paradigm to video-based multi-task learning.

\bibliographystyle{IEEEbib}
\bibliography{arxiv}
\end{document}